\documentclass[11pt,letterpaper]{article}
\usepackage{acl2017}
\usepackage{times}
\usepackage{latexsym}
\usepackage{amssymb}
\usepackage{bm}
\usepackage{graphicx}
\usepackage{lipsum}
\usepackage{booktabs}
\usepackage[para,flushleft]{threeparttable}
\usepackage{color}
\usepackage{amsmath}
\graphicspath{{\pics} }
\makeatletter
\newcommand{\ssymbol}[1]{^{\@fnsymbol{#1}}}
\makeatother

\usepackage{natbib}

\aclfinalcopy 

\title{Ruminating Reader: Reasoning with Gated Multi-Hop Attention}

\author{Yichen Gong \and Samuel R. Bowman\\
		New York University \\
        New York, NY\\
  {\tt \{yichen.gong,bowman\}@nyu.edu}}

\date{}

\begin{document}

\maketitle

\begin{abstract}
To answer the question in machine comprehension (MC) task, the models need to establish the interaction between the question and the context. To tackle the problem that the single-pass model cannot reflect on and correct its answer, we present Ruminating Reader.  Ruminating Reader adds a second pass of attention and a novel information fusion component to the Bi-Directional Attention Flow model (\textsc{BiDAF}). We propose novel layer structures that construct an query-aware context vector representation and fuse encoding representation with intermediate representation on top of \textsc{BiDAF} model. We show that a multi-hop attention mechanism can be applied to a bi-directional attention structure. In experiments on SQuAD, we find that the Reader outperforms the \textsc{BiDAF} baseline by a substantial margin, and matches or surpasses the performance of all other published systems.
\end{abstract}

\section{Introduction}

The majority of recorded human knowledge is circulated in unstructured natural language. It is tremendously valuable to allow machines to read and comprehend the text knowledge. Machine comprehension (MC)---especially in the form of question answering (QA)---is therefore attracting a significant amount of attention from the machine learning community.  Recently introduced large-scale datasets like CNN/Daily Mail \citep{CNN_DAILY_MAIL_Hermann:2015ta}, the Stanford Question Answering Dataset \citep[SQuAD;][]{Rajpurkar:2016vf} and the Microsoft MAchine Reading COmprehension Dataset \citep[MS-MARCO;][] {msmacro2016arXiv161109268N} have allow data-driven methods, including deep learning, to become viable.

\begin{figure}
\centering
\begin{tabular}{@{}c@{}}

\includegraphics[width=0.47\textwidth]{{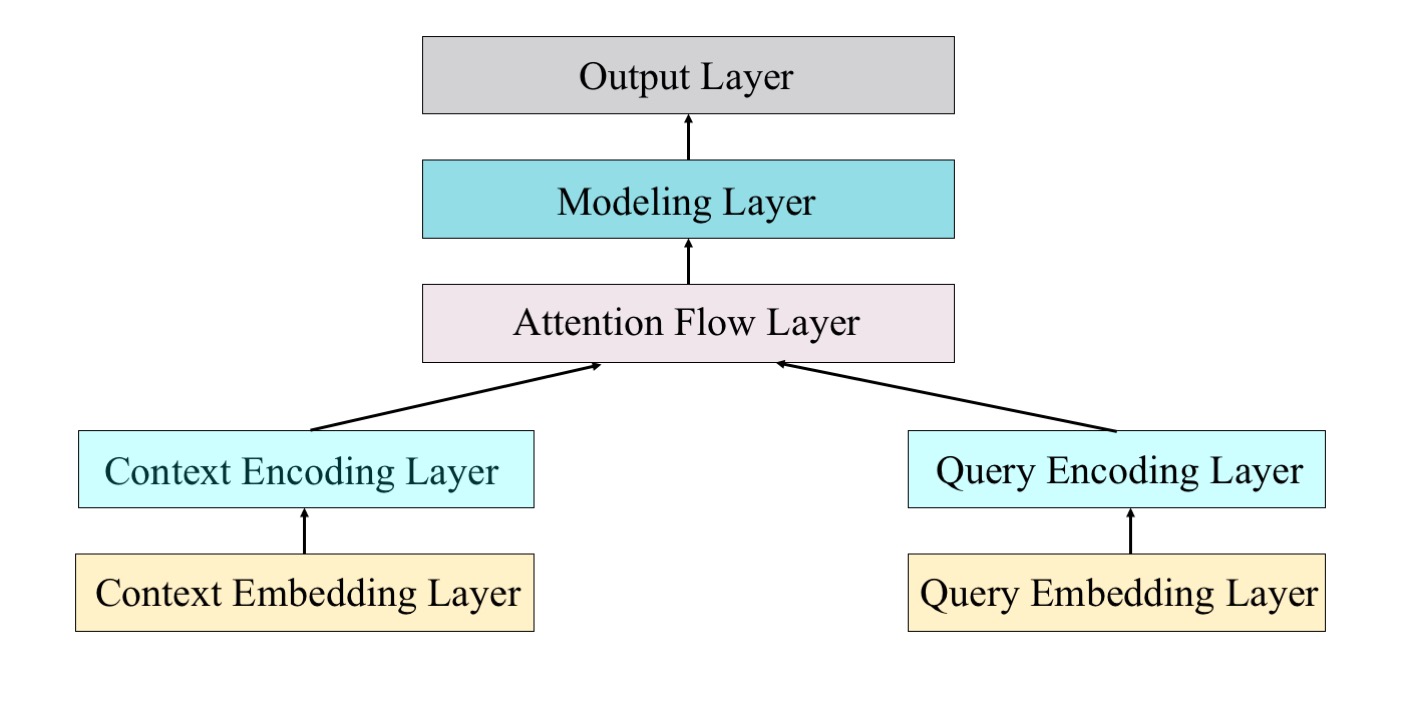}}\\
\small (a) The high-level structure of \textsc{BiDAF}.
\end{tabular}

\vspace{1em}

\begin{tabular}{@{}c@{}}
\includegraphics[width=0.47\textwidth]{{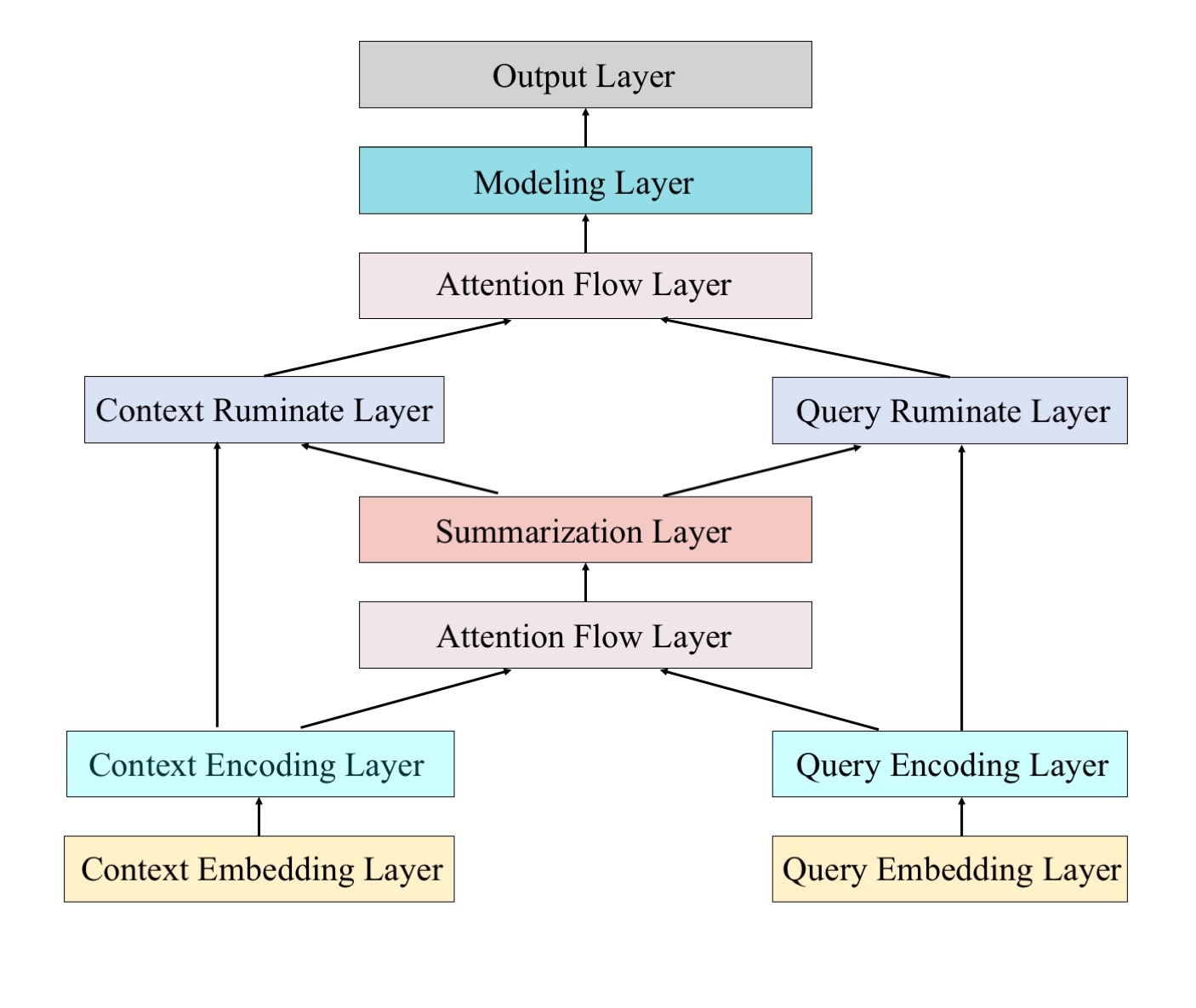}}\\
\small (b) The high-level structure of Ruminating Reader.
\end{tabular}
\caption{The high-level comparison between \textsc{BiDAF} and Ruminating Reader. } \label{fig:model_struc_simple}
\end{figure}

Recent approaches toward solving machine comprehension tasks using neural networks can be viewed as falling into two broad categories: single-pass reasoners and multiple-pass reasoners. Single-pass models read a question and a source text once and often adopt the differentiable attention mechanism that emphasizes important parts of the context related to the question.

\textsc{BiDAF} \cite{Seo:2016tp} represents one of the state-of-the-art single-pass models in Machine Comprehension. \textsc{BiDAF} uses a bi-directional attention matrix which calculates the correlations between each word pair in context and query to build query-aware context representation. However, \textsc{BiDAF} and some similar models miss some questions because they don't have the capacity to reflect on problematic candidate answers and revise their decisions.

When humans are reading a text with the goal of answering a question, they tend to read it multiple times to get a better understanding of the context and question, and to give a better response. 

With this intuition, recent multi-pass models revisit the question and the context passage (or \textit{ruminate}) to infer the relations between the context, the question and the answer. 

We propose an extension of \textsc{BiDAF}, called Ruminating Reader,  which uses a second pass of reading and reasoning to allow it to learn to avoid mistakes and to ensure that it is able to effectively use the full context when selecting an answer. In addition to adding a second pass, we also introduce two novel layer types, the \textit{ruminate layers}, which use gating mechanisms to fuse the obtained from the first and second passes. We observe a surprising phenomenon that when an LSTM layer in the context ruminate layer takes same input in each timestep, it can produce useful representation for the gates. In addition, we introduce an answer--question similarity loss to penalize overlap between question and predicted answer, a common feature in the errors of our base model. This allows us to achieve an F1 score of 79.5 and Exact Match (EM) score of 70.6 on hidden test set,\footnote{The latest results are listed at \url{https://rajpurkar.github.io/SQuAD-explorer/}} an improvement of 2.2 F1 score and 2.9 EM on \textsc{BiDAF}. Figure \ref{fig:model_struc_simple} shows a high-level comparison between \textsc{BiDAF} and Ruminating Reader.

This paper is organized as follows: In Section 2 we define the problem to be solved and introduce the SQuAD task. In Section 3 we introduce Ruminating Reader, focusing on the information-extracting and information-digesting components and how they integrate. Section 4 discusses related work. Section 5 presents the experimental setting, results and analysis. Section 6 concludes.
\section{Question Answering}

\begin{figure}[t]
\centering
\includegraphics[width=0.47\textwidth]{{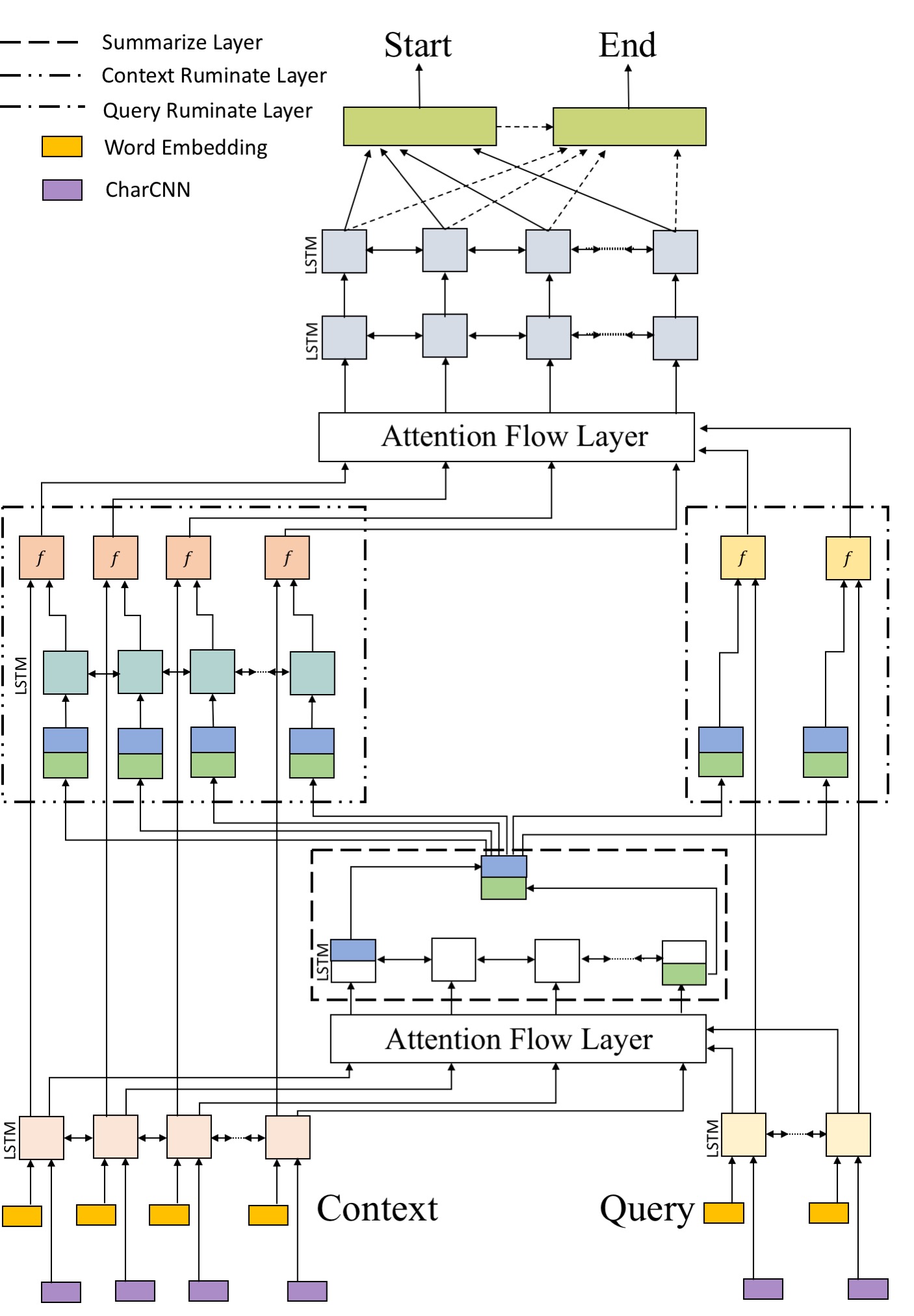}}
\caption{The model structure of our Ruminating Reader.}\label{fig:model_struc_full}
\end{figure}

The task of the Ruminate Reader is to answer a question by reading and understanding a paragraph of text and selecting a span of words within the context. Formally, the Training and development data consist of tuples (Q, P, A), where Q = $(q_1, ..., q_i, ... q_{|Q|})$ is the question, a sequence of words with length $|Q|$, C = $(c_1, ... c_j, ..., c_{|C|})$ is the context, a sequence of words with length $|C|$, and A = $(a_b, a_e)$ is the answer span marking the beginning and end indices of the the answer in the context ($1 <= a_b <= a_e <= |C|$). 

\paragraph{SQuAD}

The SQuAD corpus is built using 536 articles randomly selected from English Wikipedia. Images, figures, tables are stripped and any paragraphs shorter than 500 characters are discarded. Unlike other datasets that such as CNN/Daily Mail whose questions are synthesized, \newcite{Rajpurkar:2016vf} uses a crowdsourcing platform to generate realistic question and answer pairs. SQuAD contains 107,785 question-answer pairs. The typical context length spans from 50 tokens to 250 tokens. The typical length of a question is around 10 tokens. The answer be any span of words from the context, resulting in $O(|C|^2)$ possible outputs.

\section{Our Model}
\subsection{Ruminating Reader}

In this section, we review the \textsc{BiDAF} model \cite{Seo:2016tp} and introduce our extension, the Ruminating Reader. 

Our additions to the base model are motivated by the intuition that adding an additional pass of reading will allow the model to better integrate information from the question and answer and to better weigh possible answers, and that by interpolating the results of the second pass with those of the first pass through gating, we can prevent the additional complexity that we add to the model from substantially increasing the difficulty of training. The structure of our model is shown in Figure \ref{fig:model_struc_full} and explained in the following sections.

\paragraph{Character Embedding Layer} Just as in the base \textsc{BiDAF} model, the character embedding layer maps each word to a high dimensional vector using character features. It does so using a convolutional neural network with max pooling over learned character vectors  \cite{Fully_CharCNN_NMT_Lee:2016wl, Kim:2016vh}. Thus we have a context character representation $\bm{M}\in \mathbb{R}^{f\times C}$ and a query representation $\bm{N}\in \mathbb{R}^{f\times Q}$, where $C$ is the sequence length of the context, $Q$ is the sequence length of the query and $f$ is the number of 1D convolutional neural network filters.

\paragraph{Word Embedding Layer} Again as in the base model, the word embedding layer uses pretrained word vectors \citep[the 6B GloVe vectors of ][]{glove_Pennington:2014jd} to map the word into a high dimensional vector space. We do not update the word embeddings during training. The character embedding and the word embedding are concatenated and passed into a two-layer highway network \cite{Highway_network_2015arXiv150500387S} to obtain a $d$ dimensional vector representation of each single word. Hence, we have a context representation $\bm{H}\in \mathbb{R}^{d\times C}$ and a query representation $\bm{U}\in \mathbb{R}^{d\times Q}$.

\paragraph{Sequence Encoding Layers}  As in \textsc{BiDAF}, we use two LSTM RNNs \cite{LSTM_Hochreiter1997} with $d$-dimensional outputs to encode the context and query representations in both directions. Therefore, we obtain a context encoding matrix $\bm{C}\in \mathbb{R}^{2d \times C}$, and a query encoding matrix $\bm{Q}\in \mathbb{R}^{2d \times Q}$.

\paragraph{Attention Flow Layer} As in \textsc{BiDAF}, the attention flow layer constructs a query-aware context representation $\bm{G}$ from inputs $\bm{C}$ and $\bm{Q}$. 
This layer takes two steps. In the first step, an interaction matrix $\bm{I}\in\mathbb{R}^{C \times Q}$ is computed, which indicates the affinities between each context word encoding and each query word encoding. $\bm{I}_{cq}$ indicates the correlation between the $c$-th word in context and $q$-th word in query. The interaction matrix is computed by
\begin{equation}
\bm{I}_{cq} = \bm{w}^{\top}_{(\bm{I})}[\bm{C}_{c};\bm{Q}_{q};\bm{C}_{c}\circ \bm{Q}_{q}]
\end{equation}
where $\bm{w}_{I} \in \mathbb{R}^{6d}$ is a trainable parameter, $\bm{C}_{c}$ is $c$-th column of context encoding and $\bm{Q}_{q}$ is $q$-th column of query encoding, $\circ$ is elementwise multiplication, and $[;]$ is vector concatenation.  

\paragraph{Context-to-query Attention}  As in \textsc{BiDAF}, the context-to-query attention component generates, for each context word, an attention-weighted sum of query word encodings. Let $\bm{\tilde{Q}} \in \mathbb{R}^{2d \times C}$ represent the context-to-query attention matrix. For column $c$ in $\bm{\tilde{Q}}$ is defined by $\bm{\tilde{Q}}_{c} = \sum(\bm{a}_{cq} \bm{Q}_{q})$, where $\bm{a}$ is the attention weight. $\bm{a}$ is computed by $\bm{a}_c = softmax(\bm{I}_{c}) \in \mathbb{R}^{Q}$. 

\paragraph{Query-to-context Attention} Query-to-context attention indicates the most relevant context words to query. The most relevant word vector representation is an attention-weighted sum defined by $\bm{\tilde{c}}=\sum{\bm{b}_c\bm{C}_{c}}$ where $\bm{b}$, is an attention weight which is calculated by $\bm{b} = softmax(max_{col}(\bm{I})) \in \mathbb{R}^{C}$. $\bm{\tilde{c}}$ is replicated $C$ times across the column, therefore giving $\bm{\tilde{C}}\in \mathbb{R}^{2d \times C}$.

We then obtain the final query-aware context representation by 
\begin{equation}
\bm{G}_{c} = [\bm{C}_{c};\bm{\tilde{Q}}_{c};\bm{C}_{c}\circ \bm{\tilde{Q}}_{c};\bm{C}_{c} \circ \bm{\tilde{C}}_{c}]
\end{equation}
where $\bm{G}_{c} \in \mathbb{R}^{8d \times C}$.

\paragraph{Summarization Layer} We propose summarization layer which produces a vector representation that summarizes the information in the query-aware context representation. The input to summarization layer is $\bm{G}$. We use one bi-directional LSTM network to model the learned information. We select the final states from both directions and concatenate them together as $\bm{s}=[\bm{s}_{f};\bm{s}_{b}]$ . where $\bm{s}\in \mathbb{R}^{2d}$ represents the representation summarized from the reading of context and query, $\bm{s}_{f}$ is the final state of LSTM in forward direction, and $\bm{s}_{b}$ is the final state of LSTM in backward direction. 

\paragraph{Query Ruminate Layer} The query ruminate layer fuses the summarization vector representation with the query encoding $\bm{Q}$, helping reformulate the query representation in order to maximize the chance of retrieving the correct answer. The input to this layer is $\bm{s}$ tiled $Q$ times ($\bm{S}_Q\in \mathbb{R}^{2d\times Q}$). A gating function then fuses this with the existing query encoding:
\begin{equation}
\bm{z}_i = tanh(\bm{W}^{1\top}_{Qz} \bm{S}_{Qi} + \bm{W}^{2\top}_{Qz}\bm{Q}_i + \bm{b}_{Qz})
\end{equation}
\begin{equation}
\bm{f}_i = \sigma(\bm{W}^{1\top}_{Qf} \bm{S}_{Qi} + \bm{W}^{2\top}_{Qf}\bm{Q}_i + \bm{b}_{Qf})
\end{equation}
\begin{equation}
\tilde{\bm{Q}}_i = \bm{f}_i \circ \bm{Q}_i + (1-\bm{f}_i)\circ \bm{z}_i
\end{equation}
where $\bm{W}^1_{Qz} , \bm{W}^2_{Qz}, \bm{W}^1_{Qf}, \bm{W}^2_{Qf} \in \mathbb{R}^{2d\times 2d}$ and $\bm{b}_{Qz}, \bm{b}_{Qf} \in \mathbb{R}^{2d}$ are trainable parameters, $\bm{S}_{Qi} $is the $i$-th column of the $\bm{S}_Q$, $\bm{Q}_i$ is the $i$-th column of $\bm{Q}$.

\paragraph{Context Ruminate Layer} Context ruminate layer digests the summarization and integrates it with the context encoding $\bm{C}$ to facilitate answer extraction. In this layer, we tile $\bm{s}$ $C$ times and we have $\bm{S}_C\in \mathbb{R}^{2d\times C}$. To incorporate the positional information into this relatively long tiled sequence, we feed it into an additional bidirectional LSTM with output size $d$  in each direction. This approach, while somewhat inefficient, proves to be an valuable addition to the model and allows it to better track position information, loosely following the positional encoding strategy of \citet{memN2N_Sukhbaatar:2015ww}. Hence we obtain $\bm{\tilde{S}}_C \in \mathbb{R}^{2d \times C}$, which is fused with context encoding $\bm{C}$ via a gate:
\begin{equation}
\bm{z}_i = tanh(\bm{W}^{1\top}_{Cz} \tilde{\bm{S}}_{Ci} + \bm{W}^{2\top}_{Cz}\bm{C}_i + \bm{b}_{Cz})
\end{equation}
\begin{equation}
\bm{f}_i = \sigma(\bm{W}^{1\top}_{Cf}\tilde{\bm{S}}_{Ci} + \bm{W}^{2\top}_{Cf}\bm{C}_i + \bm{b}_{Cf} ) 
\end{equation}
\begin{equation}
\tilde{\bm{C}}_i = \bm{f}_i \circ \bm{C}_i + (1-\bm{f}_i) \circ \bm{z}_i
\end{equation}
where $\bm{W}^1_{Cz} , \bm{W}^2_{Cz}, \bm{W}^1_{Cf}, \bm{W}^2_{Cf} \in \mathbb{R}^{2d\times 2d}$ and $\bm{b}_{Cz}, \bm{b}_{Cf} \in \mathbb{R}^{2d}$ are trainable parameters, $\tilde{\bm{S}}_{Ci} $is the $i$-th column of the $\tilde{\bm{S}}_C$, $\bm{C}_i$ is the $i$-th column of $\bm{C}$.

\paragraph{Second Hop Attention Flow Layer} We take $\tilde{\bm{Q}}  \in \mathbb{R}^{2d\times Q}$ and $\tilde{\bm{C}}  \in \mathbb{R}^{2d\times C}$ as the input to another attention flow layer with the same structure as described above, yielding $\bm{G}^{(2)} \in \mathbb{R}^{8d\times C}$. 

\paragraph{Modeling Layer} We use two layers of bi-directional LSTM with output size $d$ in each direction to aggregate the information in $\bm{G}^{(2)}$, yielding a pre-output matrix $\bm{M}^s \in \mathbb{R}^{2d\times C}$. 

\paragraph{Output Layer} As in \textsc{BiDAF}, our output layer independently models the probability of each word being selected as the start or end of an answer span. We calculate the probability distribution of the start index of the answer span by 
\begin{equation}
\bm{p}^s = softmax(\bm{w}^{\top}_{\bm{p}^1}[\bm{G};\bm{M}^s])
\end{equation}
where $\bm{w}_{(\bm{p^1})}\in \mathbb{R}^{10d}$ is a trainable parameter. We pass the matrix $\bm{M}^s$ to another bi-directional LSTM with output size d in single direction yielding $\bm{M}^e$. We obtain the probability distribution of the end index of the answer span by
\begin{equation}
\bm{p}^e = softmax(\bm{w}^{\top}_{(\bm{p}^2)}[ \bm{G};\bm{M}^e])
\end{equation}

\paragraph{Training Loss} We define the training loss as the sum of three components: negative log likelihood loss,  L2 regularization loss, and a novel answer--question similarity loss.

\begin{table}
\centering
\begin{tabular}{p{0.47\textwidth}}
\toprule
\textbf{Context:}
The Broncos took an early lead in Super Bowl 50 and never
trailed. Newton was limited by Denver's defense, which
sacked him seven times and forced him into \textcolor{red}{three turnovers}, 
including \textcolor{blue}{a fumble} which they recovered for a touchdown. 
Denver linebacker Von Miller was named Super Bowl MVP, 
recording five solo tackles, 2½ sacks, and two forced 
fumbles. 
\\
\vspace{-0.1em}
\textbf{Question:}
Which Newton \textcolor{red}{turnover} resulted in seven points for Denver?\\
\vspace{-0.1em}
\textbf{Ground Truth:} \{a fumble,
a fumble,
Fumble\}\\
\vspace{-0.1em}
\textbf{Prediction:}
three turnovers\\
\bottomrule
\end{tabular}

\caption{An error of the type that motivated the answer--question similarity loss.}\label{tab:AQS_sample}
\end{table}

\paragraph{Answer--Question Similarity Loss} We observe that a version of our model trained only on the two standard loss terms often selects answers that overlap substantially in content with their corresponding questions, and that this nearly always results in an error. A sample error of this kind is shown in Table~\ref{tab:AQS_sample}. This motivates an additional loss term at training time: We penalize the similarity between the question and the selected answer. Formally, the answer question similarity loss is defined as
\begin{equation}
s = Argmax(\bm{p}^1)
\end{equation}
\begin{equation}
e = Argmax(\bm{p}^2)
\end{equation}
\begin{equation}
\bm{\vec{q}}_{BoW} = \frac{Sum_{row}(\bm{Q})}{Q}
\end{equation}

\begin{equation}
\begin{split} 
AQSL(\theta) = cos(\bm{C}_{s}, \bm{\vec{q}}_{BoW})+  cos(\bm{C}_{e}, \bm{\vec{q}}_{BoW})
\end{split}
\end{equation} 

where $s$ refers to the start index of answer span, $e$ refers to the end index of the answer span, $\bm{\vec{q}}_{BoW}$ is the bag of words representation of query encoding, $cos(a,b)$ is the cosine similarity between $a$ and $b$, $\bm{C}_s$ and $\bm{C}_e$ are the $s$-th and $e$-th vector representation of context encoding.

\paragraph{Prediction} During prediction, we use a local search strategy that for token indices $a$ and $a'$, we maximize $\bm{p}^s_{a} \times \bm{p}^e_{a'}$, where $0 \le a' - a \le 15$ . Dynamic programming is applied during search, resulting in $O(C)$ time complexity. 

\section{Related Work}
Recently, both QA and Cloze-style machine comprehension tasks like CNN/Daily Mail have seen fast progress. Much of this recent work has been based on end-to-end trained neural network models, and within that, most have used recurrent neural networks with soft attention  \cite{attention_Bahdanau:2014vz}, which emphasizes one part of the data over the others. These models can be coarsely divided into two categories: single-pass and multi-pass reasoners.

Most papers on single-pass reasoning systems propose novel ways to use the attention mechanism: \newcite{Wang:2016ws} propose match-LSTM to model the interaction between context and query, as well as introducing the use of a pointer network \cite{pointer_network_Vinyals:2015uw} to extract the answer span from the context.  \newcite{Xiong:2016uq} propose the Dynamic Coattention Network, which uses co-dependent representations of the question and the context, and iteratively updates the start and end indices to recover from local maxima and to find the optimal answer span. \newcite{Wang:2016vv} propose the Multi-Perspective Context Matching model that matches the encoded context with query by combining various matching strategies, aggregates matching vector with bi-directional LSTM, and predict start and end positions. In order to merge the entity score during its multiple appearence, \newcite{Kadlec:2016vy} propose attention-sum reader who computes dot product between context and query encoding, does a softmax operation over context and sums the probability over the same entity to favor the frequent entities over rare ones. \newcite{thorough_Chen:2016is} propose to use a bilinear term to calculate the attentional alignment between context and query.

Among multi-hop reasoning systems: \newcite{Hill:2015tgp} apply attention on window-based memory, by extending multi-hop end-to-end memory network \cite{memN2N_Sukhbaatar:2015ww}. \newcite{Gated_attention_Dhingra:2016vq} extend attention-sum reader to multi-turn reasoning with an added gating mechanism. The Iterative Alternative (IA) reader \cite{Sordoni:2016wb} produces query glimpse and document glimpse in each iterations and uses both glimpses to update recurrent state in each iteration. \newcite{Shen:2016uu} propose a multi-hop attention model that used reinforcement learning to dynamically determine when to stop digesting intermediate information and produce an answer.

\section{Evaluation}
\begin{table}[t]
\renewcommand{\thefootnote}{\fnsymbol{footnote}}
\begin{threeparttable}

\centering
\small
\begin{tabular}{l r r}
\toprule

{\bf Model} &
\multicolumn{2}{c}{\bf Test} \\

{}    & F1 & EM \\
\midrule
Logistic Regression\tnote{a} & 51.0 & 40.4\\ 
Dynamic Chunk Reader\tnote{b} & 70.956  & 62.499\\ 
Fine-grained Gating\tnote{c} & 73.327  & 62.446\\ 
Match-LSTM\tnote{d} & 73.743  & 64.744\\
Dynamic Coattention Network\tnote{e} & 75.896  & 66.233\\ 
\textbf{Bidirectional Attention Flow}\tnote{f} & 77.323  & 67.974\\
RaSoR\tnote{g} & 77.696  & 69.642\\
Multi-perspective Matching\tnote{h}& 77.771  & 68.877 \\ 
FastQAExt\tnote{i}& 78.857 & 70.849 \\
Document Reader\tnote{j} & 79.353 & 70.733\\
ReasoNet\tnote{k}& 79.364 & 70.555  \\
jNet\tnote{l}& 79.821  & 70.607 \\
Interactive AoA Reader\tnote{$\ssymbol{3}$} & 79.937 & 71.153 \\
QFASE\tnote{$\ssymbol{3}$} & 79.989 & 71.898 \\
r-net\tnote{$\ssymbol{3}$} & \textbf{80.717} & \textbf{72.338} \\
\midrule
Ruminating Reader & 79.456 & 70.639 \\
\bottomrule

\end{tabular}
\caption{The Official SQuAD leaderboard performance on test set for single model section from April 23, 2017, the time of submission. There are other unpublished systems shown on leaderboard, including Document Reader and r-net.}\label{tab:results}

\begin{tablenotes}
\item[a]\citet{Rajpurkar:2016vf};\item[b] \citet{Yu:2016vg};\item[c]\citet{Yang:2016vi};\item[d]\citet{Wang:2016ws};\item[e]\citet{Xiong:2016uq};\item[f]\citet{Seo:2016tp};\item[g]\citet{Lee:2016td};\item[h]\citet{Wang:2016vv};\item[i]\citet{Weissenborn:2017wa};\item[j]\citet{Document_reader2017arXiv170400051C}; \item[k]\citet{Shen:2016uu};\item[l]\citet{Zhang:2017wl};\item[$\ssymbol{3}$]Unpublished
\end{tablenotes}
\end{threeparttable}
\renewcommand*{\thefootnote}{\arabic{footnote}}
\end{table}

\subsection{Implementation details}

Our model configuration closely follows that of \citet{Seo:2016tp} did: In the character encoding layer, we use 100 filters of width 5. In the remainder of the model, we set the hidden layer dimension ($d$) to 100. We use pretrained 100D GloVe vectors (6B-token version) as word embeddings. Out-of-vocobulary tokens are represented by an UNK symbol in the word embedding layer, but treated normally by the character embedding layer. The BiLSTMs in context and query encoding layers share same weights. We use the AdaDelta optimizer \cite{AdaDelta2012arXiv1212.5701Z} for optimization. 

We selected hyperparameter values through random search \cite{random_grid_search_Bergstra:2012ux}. Batch size is 30. Learning rate starts at 0.5, and decreases to 0.2 once the model stops improving. The L2-regularization weight is 1e-4, AQSL weight is 1 and dropout with a drop rate of 0.2 is applied to all forward connections in the CNN, the LSTMs, and all feedforward layers. 

A typical model run converges in about 40k steps. This takes two days using Tensorflow \cite{tensorflow2015-whitepaper} and a single NVIDIA K80 GPU .

\subsection{Evaluation Method}
 \citet{Rajpurkar:2016vf} provide an official evaluation script that allows us to measure F1 score and EM score by comparing the prediction and ground truth answers. Three answers are provided for each question. The prediction is compared to each of the answer and best score is selected. F1 score is defined by recall and precision of words and EM score, as Exact Match score, is defined as the score of 100\% accuracy in prediction. We do not use any kind of ensembling, and compare our results primarily with other single-model (non-ensemble) results. The test set performance is evaluated at CodaLab by administrator.

\subsection{Results}
At the time of submission, our model is tied in accuracy on the hidden test set with the best-performing published single model \citep{Zhang:2017wl}. We achieve an F1 score of 79.5 and EM score of 70.6. The current leaderboard is displayed in Table~\ref{tab:results}. The leaderboard is listed in descending order of F1 score, but if an entry's F1 score is better than the adjacent entry's, while its EM score is worse, then these two entries are considered tied.

\begin{table}
\centering
\small
\begin{tabular}{l r r}
\toprule
{\bf Ablation Variant} &
\multicolumn{2}{c}{\bf Dev} \\

{}    & F1 & EM \\
\midrule
1. \textsc{BiDAF} & 77.3 & 67.7\\ 
2. \textsc{BiDAF} w/ L2 Reg., AQSL, LS  & 77.7 & 68.6 \\ 
3. RR w/o query ruminate layer & 78.7 & 69.6 \\ 
4. RR w/o context ruminate layer  & 78.9 & 70.0 \\
5. RR w/ BiLSTM in QRL & \textbf{79.4} & 70.4 \\
6. RR w/o BiLSTM in CRL & 74.0 & 64.2 \\
7. RR w/o query input at $\bm{s}$,$\bm{f}$ in QRL  & 78.8 & 70.1 \\
8. RR w/o context input at $\bm{s}$,$\bm{f}$ in CRL & 78.9 & 70.3 \\
9. RR w/o query input in QRL & 63.3 & 54.1 \\ 
10. RR w/o context input in CRL & 27.0 & 9.4\\
11. RR w/o summ. input in QRL & 79.2 & 70.1\\
12. RR w/o summ. input in CRL& 79.2 & 70.3 \\
\midrule
Ruminating Reader & \textbf{79.5} & \textbf{70.6} \\
\bottomrule
\end{tabular}

\caption{Layer ablation results. The order of the listing corresponds to the description in  Appendix~\ref{ssec:layer_ablation_setup}. CRL refers to context ruminate layer and QRL refers to query ruminate layer. \label{tab:Layer_Ablation_Results}}
\end{table}

\begin{figure*}[ht]
\centering
\includegraphics[width=0.98\textwidth]{{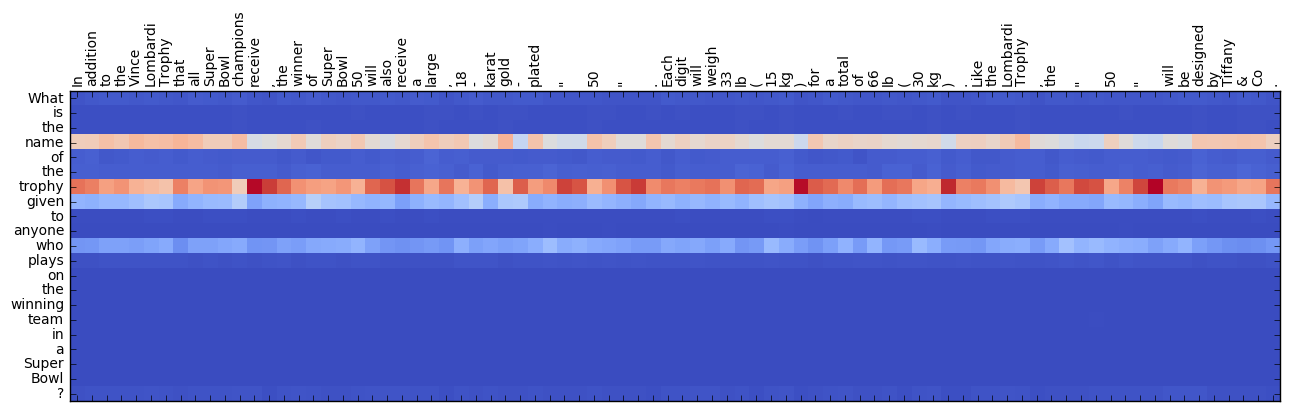}}\\
\vspace{1em}
\includegraphics[width=0.98\textwidth]{{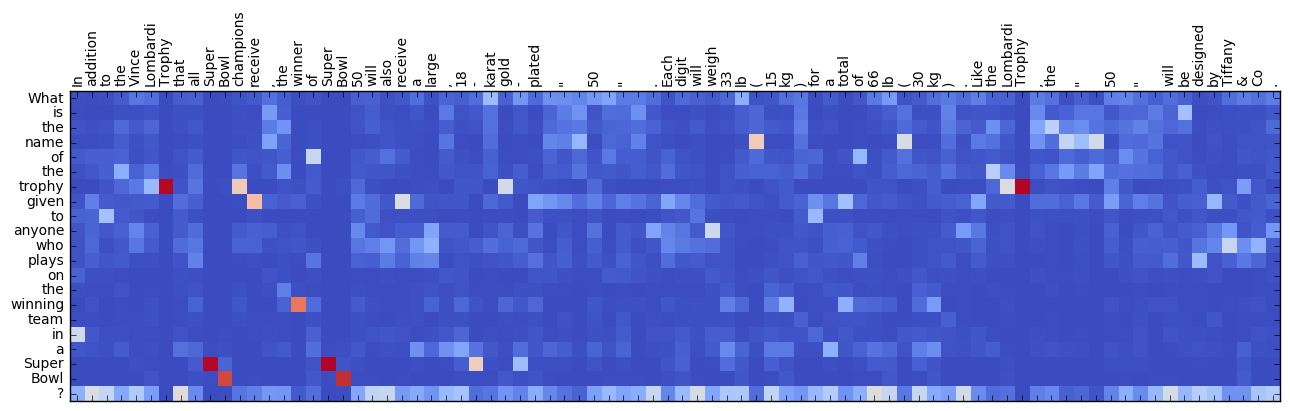}}
\caption{The visualization of first hop (top) and second hop (bottom) attention interaction matrix. We use coolwarm colormap, where red is close to 1 and blue is close to 0. In the question ``What is the name of the trophy given to anyone who plays on the winning team in a super Bowl?'', the key words \textit{name}, \textit{trophy}, \textit{given}, \textit{who} are strongly attended to in the first hop.} \label{fig:first_hop_att}
\end{figure*}

\subsection{Analysis}

\paragraph{Layer Ablation Analysis} To analyze how each component contribute to the model, we run a layer ablation experiment. We present results for twelve versions of the model on the development set, each missing some or all of the major components of the full Ruminating Reader. The precise definition of each of the twelve ablated models can be found in Appendix \ref{ssec:layer_ablation_setup}.

The results of the ablation experiment are shown in Table \ref{tab:Layer_Ablation_Results}. The ablation experiments show how each component contribute to the model. Experiments 3 and 4 show that the two ruminate layers are both important and helpful in contributing performance. It is worth noting that the BiLSTM in the context ruminate layer contributes substantially to model performance. We find this somewhat surprising, since it takes the same input in each timestep, but it nonetheless successfully digests the summarization information representation and produces a useful input for the gating component. Experiments 7 and 8 show that the modeled summarization vector representation can provide information to gates reasonably well. The drop in performance in both experiments 9 and 10 shows that the key information for new query and context representation are the are first stage query and context encodings. Experiments 11 and 12 shows that the summarization vector representation does help the later stage of reasoning.

\begin{figure}[ht]
\centering
\includegraphics[width=0.47\textwidth]{{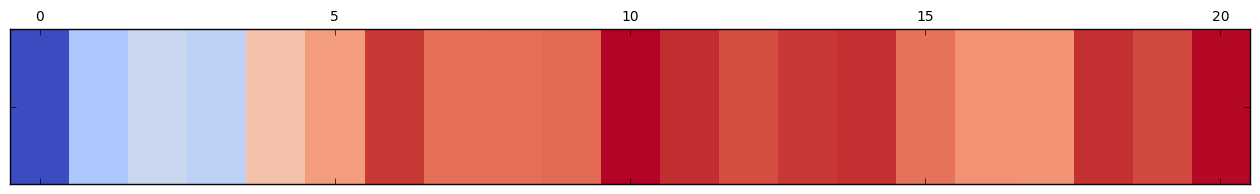}}
\\\vspace{1em}
\includegraphics[width=0.47\textwidth]{{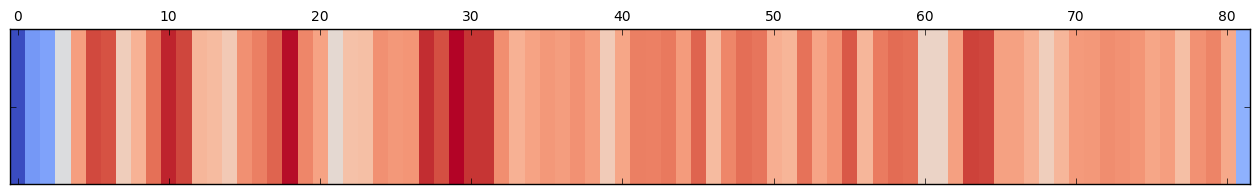}}
\caption{Visualizations of gate values for the query (top) and context (bottom) ruminate layers. The order of words is the same as in Figure~\ref{fig:first_hop_att}. We use coolwarm colormap, where red means the gate uses more information from intermediate representation, and blue from encoding representation.} \label{fig:context_gate_vis}
\end{figure}

\begin{figure}[t]
\centering
\includegraphics[width=0.47\textwidth]{{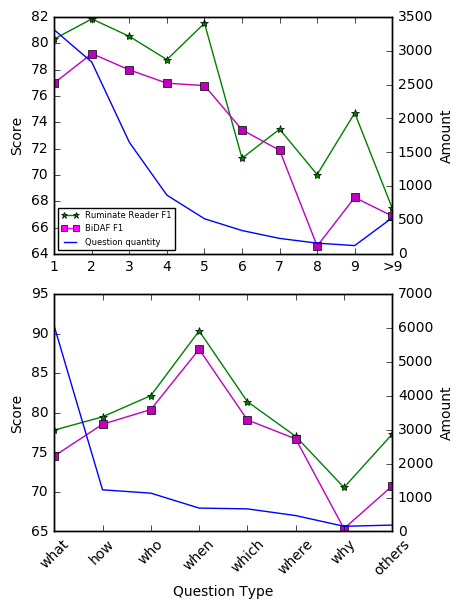}}
\caption{An analysis according to answer length of questions and question types. The top graph shows the comparison of F1 score between Ruminating Reader and the \textsc{BiDAF} model on the development set by answer length. The blue line shows the distribution of questions by answer length. The bottom graph shows a corresponding comparison \textsc{BiDAF} by question type (\textit{wh}-word).} \label{fig:F1_vis}
\end{figure}

\paragraph{Visualization} Figure~\ref{fig:first_hop_att} provides a visualization of the first hop and second hop attention interaction matrix $\bm{I}$. We also provide a sample of visualization for the L2 sum of gate value in context and query ruminate layers in Figure~\ref{fig:context_gate_vis}.

From Figure~\ref{fig:first_hop_att} we see that though the structures of two hops of attention flow layer are the same, they function quite differently in typical cases. The first hop attention appears to be primarily concerned with identifying the key informative word (or words, as here) in the query. Though in Figure~\ref{fig:first_hop_att} four key words are signified, one or two words are attended to in the first hop in the common case. The second hop is then responsible for finding candidate answers that are relevant to those key words and generating a query-aware context representation. We observe the first hop attention shows a consistent attention pattern across context words, suggesting that there may be room to make the first hop component more efficient in future work.

From Figure~\ref{fig:context_gate_vis}, we see the gate value on both query ruminate layer and context ruminate layer shows that the gates are working to fuse information to original query encoding and context encoding. We observe that in most of the case the gates in ruminate layers uses more information from encoding than from summarization representation. The observation matches our expectation that the gates modify and improve on the encoding representation.

We also provide a comparison of F1 score between \textsc{BiDAF} and Ruminating Reader on question with different ground truth answer length and different types of questions in Figure~\ref{fig:F1_vis}. Exact match score is highly correlated with F1 score so we omit it for clarity. We observe that the Ruminating Reader outperforms \textsc{BiDAF} on most of the questions with respect of different answer length. On the question with long answer length, of 5, 8 and 9, Ruminating Reader outperforms \textsc{BiDAF} by a great margin. Questions with longer reference answers appear to be more difficult to answer. In addition, the Ruminating Reader does better on each type of question. Both models work best for \textit{when} questions---these question are answerable by temporal expressions, which are relatively easy to recognize. The \textit{Why} questions are hardest to answer---they tend to have long answers with no purely lexical cues marking their beginnings or ends. Ruminating Reader outperforms \textsc{BiDAF} model on \textit{why} questions by a substantial margin.

\noindent{\textbf{Performance Breakdown}}  Following \citet{Zhang:2017wl}, we break down Ruminating Reader's 79.5\% F1 score on the development set into three sub-scores, representing failures, partial successes, and successes. On 13.5\% of development set examples, Ruminate Reader fails, yielding 0\% F1. On 70.6\% of examples, Ruminate Reader achieves a perfect F1 score. On the remaining 15.9\%, Ruminate Reader got only partial matches (i.e., answers that partially overlapped with reference answers), with an average F1 score of 56.0\%. Comparing to the jNet \cite{Zhang:2017wl} whose success answers occupy 69.1\% of all answers, failure score answers 14.9\% and partial success 16.01\% with an average F1 score of 58.0\%, our model works better on increasing successes and reducing failures.

\section{Conclusion}
We propose the Ruminating Reader, an extension to the \textsc{BiDAF} model with two-hop attention. The model surpasses the original \textsc{BiDAF}
model's performance on Stanford Question Answering Dataset (SQuAD) by a large margin, and ties with the best published system. These results and our qualitative analysis both suggest that the model successfully fuses the information from two passes of reading using gating and uses the result to identify appropriate answers to Wikipedia questions. An ablation experiment shows that each of components of this complex model contribute substantially. In future work, we aim to find ways to simplify this model without impacting performance, to explore the possibility of yet deeper models, and to expand our study to machine comprehension tasks more broadly.

\section*{Acknowledgments}
We thank Pranav Rajpurkar for testing Ruminate Reader on SQuAD hidden test set.

\bibliography{acl2017}
\bibliographystyle{acl_natbib}
\appendix
\section{Appendix}

\subsection{Layer Ablation Experiments setup} \label{ssec:layer_ablation_setup}
In this section we show the setup of layer ablation experiment details in Table \ref{tab:Layer_Ablation_Results}. 

\begin{enumerate}
\item  Vanilla \textsc{BiDAF}
\item  Ruminating Reader without context and query ruminate layer. Therefore, the model is equivalent to original \textsc{BiDAF} model with L2-regurization, answer-question similarity penalization and local search prediction feature.
\item  Ruminating Reader without query ruminate layer. The query encoding $\bm{Q}$ is directly fed into the second hop attention flow layer. 
\item  Ruminating Reader without context ruminate layer. The context encoding $\bm{C}$ is directly connected to the second hop attention flow layer without digesting newly acquired information. 

\item  Ruminating Reader with BiLSTM modeling in query ruminate layer. Formally, we have $\tilde{\bm{S}}_Q = BiLSTM(\bm{S}_Q)$ in query ruminate layer. Therefore, the query ruminate layer is defined by
\begin{equation}
\bm{z}_i = tanh(\bm{W}^{1\top}_{Qz} \tilde{\bm{S}}_{Qi} + \bm{W}^{2\top}_{Qz}\bm{Q}_i + \bm{b}_{Qz})
\end{equation}
\begin{equation}
\bm{f}_i = \sigma(\bm{W}^{1\top}_{Qf} \tilde{\bm{S}}_{Qi} + \bm{W}^{2\top}_{Qf}\bm{Q}_i + \bm{b}_{Qf})
\end{equation}
\begin{equation}
\tilde{\bm{Q}}_i = \bm{f}_i \circ \bm{Q}_i + (1-\bm{f}_i)\circ \bm{z}_i
\end{equation}

\item  Ruminating Reader without BiLSTM modeling in context ruminate layer. Formally, we have $\tilde{\bm{S}}_C = \bm{S}_C$ in query ruminate layer and all other components remains the same.
\item  Ruminating Reader without query input at $\bm{z}$, $\bm{f}$ in query ruminate layer. While all other components remain the same as in Ruminating Reader, the gate in query ruminate layer is defined by
\begin{equation}
\bm{z}_i = tanh(\bm{W}^{1\top}_{Qz} \bm{S}_{Qi} + \bm{b}_{Qz})
\end{equation}
\begin{equation}
\bm{f}_i = \sigma(\bm{W}^{1\top}_{Qf}\bm{S}_{Qi} + \bm{b}_{Qf})
\end{equation}
\begin{equation}
\tilde{\bm{Q}}_i = \bm{f}_i \circ \bm{Q}_i + (1-\bm{f}_i)\circ \bm{z}_i
\end{equation}

\item Ruminating Reader without context input at  $\bm{z}$, $\bm{f}$ in context ruminate layer. While all other components  remain the same as in Ruminating Reader, the gate in context ruminate layer is defined by
\begin{equation}
\bm{z}_i = tanh(\bm{W}^{1\top}_{Cz} \tilde{\bm{S}}_{Ci}+ \bm{b}_{Cz})
\end{equation}
\begin{equation}
\bm{f}_i = \sigma(\bm{W}^{1\top}_{f}\tilde{\bm{S}}_{Ci} + \bm{b}_{Cf} )
\end{equation}
\begin{equation}
\tilde{\bm{C}}_i = \bm{f}_i \circ \bm{C}_i + (1-\bm{f}_i) \circ \bm{z}_i
\end{equation}

\item Ruminating Reader without query input in query ruminate layer. In this version, we discard query encoding input $\bm{Q}$ in the gate of query ruminate layer. Formally, the gate in Query Ruminate layer is 

\begin{equation}
\bm{z}_i = tanh(\bm{W}^{1\top}_{Qz} \bm{S}_{Qi} + \bm{b}_{Qz})
\end{equation}
\begin{equation}
\bm{f}_i = \sigma(\bm{W}^{1\top}_{Qf} \bm{S}_{Qi} + \bm{b}_{Qf})
\end{equation}
\begin{equation}
\tilde{\bm{Q}}_i = (1-\bm{f}_i)\circ \bm{z}_i
\end{equation}

\item Ruminating Reader without context encoding input in context ruminate layer. We ablate the context encoding input $\bm{C}$ in the gate of context ruminate layer. Therefore, the gate in context ruminate layer is
\begin{equation}
\bm{z}_i = tanh(\bm{W}^{1\top}_{Cz} \tilde{\bm{S}}_{Ci} + \bm{b}_{Cz})
\end{equation}
\begin{equation}
\bm{f}_i = \sigma(\bm{W}^{1\top}_{f}\tilde{\bm{S}}_{Ci} + \bm{b}_{Cf} )
\end{equation}
\begin{equation}
\tilde{\bm{C}}_i = (1-\bm{f}_i) \circ \bm{z}_i
\end{equation}

\item Ruminating Reader without summarization information input in query ruminate layer. In case that the summarization do not help the encoding, while on the other hand, the gate contributes to the learning, we design the experiment that allows to eliminate the influence of summarization. We discard the summarization input in query ruminate layer. Formally, the gate in query ruminate layer is defined as
\begin{equation}
\bm{z}_i = tanh(\bm{W}^{2\top}_{Qz}\bm{Q}_i + \bm{b}_{Qz}  )
\end{equation}
\begin{equation}
\bm{f}_i = \sigma(\bm{W}^{2\top}_{Qf}\bm{Q}_i + \bm{b}_{Qf})
\end{equation}
\begin{equation}
\tilde{\bm{Q}}_i = \bm{f}_i \circ \bm{Q}_i + (1-\bm{f}_i)\circ \bm{z}_i
\end{equation}

\item Ruminating Reader without summarization information input in context ruminate layer. The summarization information is not included in $\bm{z}_i, \bm{f}_i$

\begin{equation}
\bm{z}_i = tanh(\bm{W}^{2\top}_{Cz}\bm{C}_i + \bm{b}_{Cz})
\end{equation}
\begin{equation}
\bm{f}_i = \sigma(\bm{W}^{2\top}_{Cf}\bm{C}_i + \bm{b}_{Cf})
\end{equation}
\begin{equation}
\tilde{\bm{C}}_i = \bm{f}_i \circ \bm{C}_i + (1-\bm{f}_i) \circ \bm{z}_i
\end{equation}

\end{enumerate}

\end{document}